\documentclass[journal]{IEEEtran}
\ifCLASSINFOpdf
\else
\fi
\usepackage{amsmath}

\usepackage{amssymb,xspace,intmacros,subfigure,amsfonts}

\usepackage[final]{graphicx}

\newcommand{\xb}{{\textbf{x}}}
\newcommand{\yb}{{\textbf{y}}}
\newcommand{\Yb}{{\textbf{Y}}}
\newcommand{\sbb}{{\textbf{s}}}
\newcommand{\Sbb}{{\textbf{S}}}
\newcommand{\Xb}{{\textbf{X}}}
\newcommand{\db}{{\textbf{d}}}
\newcommand{\Db}{{\textbf{D}}}

\newcommand{\Ab}{{\textbf{A}}}
\newcommand{\nb}{{\textbf{n}}}

\newcommand{\vb}{{\textbf{v}}}
\newcommand{\Vb}{{\textbf{V}}}

\newcommand{\lone}{$\ell^{1}$}
\newcommand{\ltwo}{$\ell^{2}$}

\newcommand\ie{i.e.\xspace}

\begin{document}
\title{Dictionary Learning for Blind One Bit Compressed Sensing}

\author{Hadi~Zayyani,
Mehdi~Korki,~\IEEEmembership{Student Member,~IEEE,}
        and~Farrokh~Marvasti,~\IEEEmembership{Senior Member,~IEEE}% <-this % stops a space
\thanks{H. Zayyani is with the Department
of Electrical and Computer Engineering, Qom University of Technology, Qom, Iran (e-mail: zayyani2009@gmail.com).}% <-this %
\thanks{M. Korki is with the Department
of Telecommunications, Electrical, Robotics and Biomedical Engineering, Swinburne University of Technology, Hawthorn,
3122 Australia (e-mail: mkorki@swin.edu.au).}
\thanks{F. Marvasti is with the Department
of Electrical Engineering, Sharif University of Technology, Tehran, Iran (e-mail: marvasti@sharif.edu).}

%stops a space
%\thanks{Manuscript received April 19, 2005; revised December 27, 2012.}
}


\markboth{IEEE Signal Processing Letters,~Vol.XX, No.X}%
{Shell \MakeLowercase{\textit{et al.}}:}



\maketitle

\begin{abstract}
This letter proposes a dictionary learning algorithm for blind one bit compressed sensing. In the blind one bit compressed sensing framework, the original signal to be reconstructed from one bit linear random measurements is sparse in an unknown domain. In this context, the multiplication of measurement matrix $\Ab$ and sparse domain matrix $\Phi$, \ie $\Db=\Ab\Phi$, should be learned. Hence, we use dictionary learning to train this matrix. Towards that end, an appropriate continuous convex cost function is suggested for one bit compressed sensing and a simple steepest-descent method is exploited to learn the rows of the matrix $\Db$. Experimental results show the effectiveness of the proposed algorithm against the case of no dictionary learning, specially with increasing the number of training signals and the number of sign measurements.

\end{abstract}


 \ifCLASSOPTIONpeerreview
 \begin{center} \bfseries EDICS: SAS-ADAP \end{center}
 \fi
\IEEEpeerreviewmaketitle

\section{Introduction}

\IEEEPARstart{T}{he} one bit compressed sensing which is the extreme case of quantized compressed sensing \cite{ZymnBC10} has been extensively investigated recently \cite{BoufB08}-\cite{Chen15}. According to compressed sensing (CS) theory, a sparse signal can be reconstructed from a number of linear measurements which could be much smaller than the signal dimension \cite{CandT06},~\cite{Dono06}. Classical CS neglects the quantization process and assumes that the measurements are real continuous valued. However, in practice the measurements should be quantized to some discrete levels. This is known as quantized compressed sensing \cite{ZymnBC10}. In the extreme case, there are only two discrete levels. This is called one bit compressed sensing and it has gained much attention in the research community recently \cite{BoufB08}-\cite{Chen15}, specially in wireless sensor networks \cite{Chen15}. In the one bit compressed sensing framework, it is proved that an accurate and stable recovery can be achieved by using only the sign of linear measurements \cite{JacqLBB13}.

Many algorithms have been developed for one bit compressed sensing. A renormalized fixed-point iteration (RFPI) algorithm which is based on \lone-norm minimization has been presented in \cite{BoufB08}. Also, a matching sign pursuit (MSP) algorithm has been proposed in \cite{Bouf09}. A binary iterative hard thresholding (BIHT) algorithm introduced in \cite{JacqLBB13}, which has been shown to have better recovery performance than that of MSP. Moreover, a restricted-step shrinkage (RSS) algorithm which has been devised in \cite{Lask11} has provable convergence guarantees.

 In addition to noise-free settings, there may be noisy sign measurements. In this case, we may be encountered with sign flips which will worsen the performance. In \cite{YanYO12}, an adaptive outlier pursuit (AOP) algorithm is developed to detect the sign flips and reconstruct the signals with very high accuracy even when there are a large number of sign flips \cite{YanYO12}. Moreover, noise-adaptive RFPI (NARFPI) algorithm combines the idea of RFPI and AOP \cite{MovaPD12}. In addition, \cite{PlanV13} proposes a convex approach to solve the problem. Recently, a one bit Bayesian compressed sensing \cite{Li15} and a MAP approach \cite{DongZ15} have been developed for solving the problem.

The basic assumption imposed by CS is that the signal is sparse in a domain, \ie in a dictionary. The dictionary is a predefined dictionary or it may be constructed for a class of signals. The dictionary learning algorithm attempts to find an adaptive dictionary for sparse representation of a class of signals \cite{Fros11}. The most important dictionary learning algorithms are the method of optimal directions (MOD) \cite{Enga99} and K-SVD \cite{AharE06}. There are some research work that use dictionary learning algorithm in the CS framework \cite{Carv09}, \cite{Chen13}, \cite{Chen15}. However, to the best of our knowledge, investigation of the dictionary learning algorithm in the one bit CS framework has not been reported in the literature.

In this letter, similar to blind CS \cite{GleiE11}, we assume that the sparse domain is unknown in advance. In conventional one bit CS, we need to know both the measurement matrix $\Ab$ and sparse domain matrix $\Phi$ to form a multiplication matrix $\Db=\Ab\Phi$. However, in the sequel we assume that the sparse domain matrix $\Phi$ is unknown. Thus, we learn the matrix $\Db$ by minimizing an appropriate cost function. The proposed algorithm similar to the most of the dictionary learning algorithms iterates between two steps. The first step is the one bit CS and the second step is the dictionary update step. Simulation results show the effectiveness of the proposed algorithm for reconstructing the sparse vector from one bit linear random measurements, specially when the number of training signals and sign measurements is large.

The rest of the letter is organized as follows. Section~\ref{sec: Alg} introduces our proposed algorithm, including problem formulation and the two steps of the algorithm. Simulation results are presented in Section~\ref{sec: Sim}. Finally, conclusions are drawn in Section~\ref{sec: con}.

\section{The proposed algorithm}
\label{sec: Alg}
\subsection{Problem Formulation}
Consider a signal $\xb_i=\Phi\sbb_i$ in an unknown domain $\Phi \in \mathbb{R}^{m \times K}$, where $\sbb_i \in \mathbb{R}^{K}$ is a sparse vector. In one bit compressed sensing, only the sign of the linear random measurements are available, i.e,
\begin{equation}
\yb_i=\mathrm{sign}(\Ab\xb_i+\vb_i),
\end{equation}
where $\Ab \in \mathbb{R}^{n \times m}$ is a random measurement matrix, $\yb_i \in \mathbb{Z}^{n}$ is the sign measurement vector and $\vb_i \in \mathbb{R}^{n}$ is the noise measurement vector, which is assumed to be i.i.d random Gaussian with variance $\sigma_n^2$. We aim to estimate the sparse vector $\sbb_i$ from only sign measurements $\yb_i$. The problem is to find $\sbb_i$ and then $\xb_i$ from the sign measurements
\begin{equation}
\yb_i=\mathrm{sign}(\Db\sbb_i+\vb_i),
\end{equation}
where hereafter, the matrix $\Db=\Ab\Phi$ is called dictionary. The sparse domain $\Phi$ is unknown in advance. As a result, the dictionary $\Db$ is also unknown. We use some training signals $\yb_i, 1\le i\le T$  to learn the dictionary matrix $\Db$ from sign measurements, where $T$ is the number of training signals. The overall problem of dictionary learning for one bit compressed sensing is to find the sparse matrix $\Sbb=[\sbb_1|\sbb_2|...|\sbb_T]$ and then $\Xb=[\xb_1|\xb_2|...|\xb_T]$ from a training matrix $\Yb=[\yb_1|\yb_2|...|\yb_T]$ which is
\begin{equation}
\Yb=\mathrm{sign}(\Db\Sbb+\Vb),
\end{equation}
where $\Vb=[\vb_1|\vb_2|...|\vb_T]$ is the collection of noise vectors. After learning the matrix $\Db$ from the proposed algorithm and finding the sparse matrix $\hat{\Sbb}$, the estimate of matrix $\Phi$ is $\hat{\Phi}=\Ab_{\mathrm{left}}^{-1}\Db$ where $\Ab_{\mathrm{left}}^{-1}=(\Ab^T\Ab)^{-1}\Ab^T$ is the left inverse matrix of $\Ab$. Finally, the estimate of the original signals will be $\hat{\Xb}=\hat{\Phi}\hat{\Sbb}$.

\subsection{Two steps of the proposed algorithm}
Inspired by the most of dictionary learning algorithms \cite{Fros11}, we divide the problem into two steps. The first step is the sparse recovery from one bit measurements when the dictionary is fixed. The second step is the dictionary update when the sparse coefficients are fixed.
\subsubsection{One bit compressed sensing: Dictionary $\Db$ is fixed}
Various algorithms were proposed to solve the conventional one bit compressed sensing (CS) problem, such as BIHT \cite{JacqLBB13}, MSP \cite{Bouf09}, RFPI \cite{BoufB08}, and  AOP \cite{YanYO12}, to name a few. Because of its simplicity, in this letter, we use BIHT algorithm to perform sparse recovery for all of the training signals. Note that the proposed dictionary learning algorithm can use any of the sparse recovery methods in the one bit CS framework. For notational convenience, we use the following notation for this step:
\begin{equation}
\label{eq: onebit}
\hat{\sbb}_i=\mathrm{BIHT}(\yb_i,\Db)\quad 1\le i\le T.
\end{equation}

\subsubsection{Dictionary update: Sparse matrix $\Sbb$ is fixed}
For dictionary update, since we have only the sign of measurements, it is infeasible to use the classical dictionary learning algorithms such as MOD \cite{Enga99} or K-SVD \cite{AharE06}. In order to learn the dictionary, we propose the following cost function for the one bit CS framework:
\begin{equation}
\label{eq:cd}
C(\Db)=\sum_{i=1}^T\sum_{k=1}^n \mathrm{Ind}(y_{ik}-\mathrm{sign}(\db_k^T\sbb_i+\nb_i)),
\end{equation}
where $\db_k^T$ is the $k$'th row of the dictionary matrix $\Db$ and $\mathrm{Ind}(x)$ is the indicator function which is defined as:
\begin{equation}
\mathrm{Ind}(x)=\Big\{\begin{array}{cc}
                    0 & x=0, \\
                    \infty & x\neq 0.
                  \end{array}
\end{equation}
Therefore, the dictionary update step is to solve the following optimization problem
\begin{equation}
\label{eq:P1}
\underset{\Db \in \mathbb{R}^{n \times K}}{\text{minimize}}\quad
C(\Db),
%\mathrm{Min}_{\Db} \quad\sum_{i=1}^T\sum_{k=1}^n \mathrm{Ind}(y_{ik}-\mathrm{sign}(\db_k^T\sbb_i+\nb_i)).
\end{equation}
where $C(\Db)$ is given in (\ref{eq:cd}).
The optimization problem in (\ref{eq:P1}) can be divided into $n$ sub-optimization problems to find the rows ($\db_k^T$) of the dictionary matrix $\Db$. The sub-optimization problems are
\begin{equation}
\label{eq: opt}
\underset{\db_k \in \mathbb{R}^{n}}{\text{minimize}}\quad\sum_{i=1}^T \mathrm{Ind}(y_{ik}-\mathrm{sign}(\db_k^T\sbb_i+\nb_i)),\quad 1\le k\le n.
\end{equation}
To solve (\ref{eq: opt}), we use two continuous approximations of the two functions $\mathrm{Ind}(x)$ and $\mathrm{sign}(x)$. For sign function, we use a continuous S-shaped function as
\begin{equation}
\label{eq: S}
\mathrm{sign}(x)\approx \mathrm{S}(x)=\frac{1-\exp(-x)}{1+\exp(-x)}.
\end{equation}
For indicator function, inspired by the definition of \lone-norm and \ltwo-norm, we define two indicator functions:
\begin{equation}
\mathrm{Ind}(x)=I(x)=\Big\{\begin{array}{cc}
                    |x| & L1 \quad\mathrm{indicator}\quad\mathrm{function}, \\
                    x^2 & L2 \quad\mathrm{indicator}\quad\mathrm{function}.
                  \end{array}
\end{equation}
Hence, the sub-optimization problem is
\begin{equation}
\label{eq: opt1}
\underset{\db_k \in \mathbb{R}^{n}}{\text{minimize}}\quad F(\db_k)=\sum_{i=1}^T I(y_{ik}-\mathrm{S}(\db_k^T\sbb_i+\nb_i)),\quad 1\le k\le n.
\end{equation}
Thanks to the approximations, the cost function $F(\db_k)$ in (\ref{eq: opt1}) is a continuous cost function. It can be shown that with neglecting the noise term and considering the \ltwo-norm and \lone-norm indicator functions, the deterministic cost functions $J(\Db)=\sum_{i=1}^T||\yb_i-\mathrm{S}(\Db\sbb_i)||_2^2$ and $Q(\Db)=\sum_{i=1}^T||\yb_i-\mathrm{S}(\Db\sbb_i)||_1$  are convex with respect to $\Db$. The proof is postponed to Appendix \ref{sec: app1} and Appendix \ref{sec: app2}, respectively. Hence, both have unique minimizer $\Db_{\mathrm{opt1}}$ and $\Db_{\mathrm{opt2}}$ which can be found by $n$ parallel simple steepest-descent methods, each responsible for finding a row of the dictionary $\Db$. The recursion of the $k$'th steepest-descent is
\begin{equation}
\label{eq: sd0}
\db_k=\db_k-\mu\frac{\partial F(\db_k)}{\partial \db_k},
\end{equation}
with the partial derivative
\begin{equation}
\label{eq: sd}
\frac{\partial F(\db_k)}{\partial \db_k}=\sum_{i=1}^T I'(y_{ik}-\mathrm{S}(\db_k^T\sbb_i))\frac{\partial}{\partial \db_k}(-\mathrm{S}(\db_k^T\sbb_i)),
\end{equation}
where $I'(x)$ is the derivative of function $I(x)$. Substituting (\ref{eq: sd}) into (\ref{eq: sd0}) results in the following final recursion
\begin{equation}
\label{eq: finalsd}
\db_k=\db_k+\mu\sum_{i=1}^T \sbb_i\mathrm{S}'(\db_k^T\sbb_i)I'(y_{ik}-\mathrm{S}(\db_k^T\sbb_i)),
\end{equation}
where $\mathrm{S}'(x)=\frac{2\exp(-x)}{1+\exp(-x)}$ and $\mu$ are the derivative of $\mathrm{S}(x)$ and the step-size parameter, respectively. In the case of \ltwo-norm indicator function with $I(x)=x^2$, the final recursion is
\begin{equation}
\label{eq: finalsd1}
\db_k=\db_k+\mu\sum_{i=1}^T \sbb_i\mathrm{S}'(\db_k^T\sbb_i)e_{ik},
\end{equation}
where $e_{ik}=y_{ik}-\mathrm{S}(\db_k^T\sbb_i)$. Regarding \lone-norm indicator function with $I(x)=|x|$, the steepest-descent recursion is
\begin{equation}
\label{eq: finalsd2}
\db_k=\db_k+\mu\sum_{i=1}^T \sbb_i\mathrm{S}'(\db_k^T\sbb_i)\mathrm{sign}(e_{ik}).
\end{equation}
Therefore, the overall algorithm is a two-step iterative algorithm which iterates either between (\ref{eq: onebit}) and (\ref{eq: finalsd1}) in the case of \ltwo-norm indicator function or between (\ref{eq: onebit}) and (\ref{eq: finalsd2}) regarding \lone-norm indicator function. We call these two versions of our algorithm DL-BIHT-L2 and DL-BIHT-L1, respectively.

\section{Simulation Results}
\label{sec: Sim}
This section presents the simulation results. In the simulations, the unknown sparse vector $\sbb_i$ is drawn from a Bernoulli Gaussian (BG) model with activity probability $p=0.01$ and with variance of active samples $\sigma^2_r=1$. To resolve the amplitude ambiguity arisen in one bit compressed sensing, we normalized the sparse vector $\sbb_i$ to have unit norm. The size of the signal vector $\xb_i$ is assumed to be $m=50$. The elements of sensing matrix $\Ab$ are obtained from a standard Gaussian distribution with $a_{ij}\sim\mathcal{N}(0,1)$. The elements of sparse domain matrix  $\Phi$ are assumed to be drawn from a standard Gaussian distribution. The columns of this matrix are also normalized to have unit norm. The number of atoms are assumed to be $K=100$. The additive noise $\vb_i$ is considered as Gaussian random variable with distribution $v_{ki}\sim\mathcal{N}(0,\sigma^2_n)$ where $\sigma_n=0.01$. For initialization of the dictionary $\Db$, we use a perturbed version of $\Db$ which is $\Db_{\mathrm{init}}=\Db+0.1\times \mathrm{randn}(n,K)$. For the BIHT algorithm, we used 20 iterations with the parameter $\tau=1$ \cite{JacqLBB13}.

\begin{figure}[tb]
\begin{center}
\includegraphics[width=8cm]{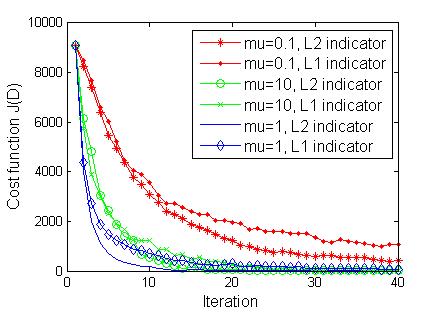}
\end{center}
\caption{Cost function versus the number of iterations.}
%\end{center}
\label{fig1}
\end{figure}

In the first experiment, we examine the convergence behavior of the proposed cost function for different values of step-size parameter $\mu$. The number of iterations is selected as 40 which is sufficient for the convergence of the cost function in most of the simulation cases. The number of training signals is $T=100$. The number of sign measurements is assumed to be $n=100$. Figure~\ref{fig1} shows the cost function $J(\Db)=\sum_{i=1}^T||\yb_i-\mathrm{S}(\Db\sbb_i)||_2^2$ versus the number of iterations for both DL-BIHT-L2 and DL-BIHT-L1 and for three values of $\mu=0.1$, $\mu=1$ and $\mu=10$. It is seen that both DL-BIHT-L2 and DL-BIHT-L1 exhibit a monotone decreasing cost functions that achieve the lowest values after almost 40 iterations. Among the three values for step size $\mu$, the best value is $\mu=1$ which leads to the fastest convergence. We use this value in the next experiments.

\begin{figure}[tb]
\begin{center}
\includegraphics[width=8cm]{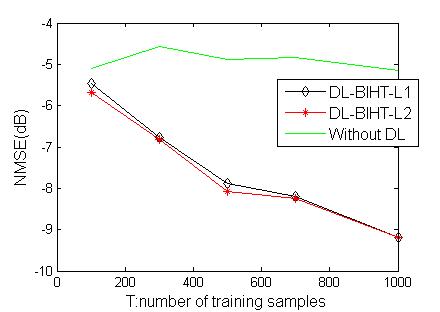}
\end{center}
\caption{NMSE versus number of training signals.}
%\end{center}
\label{fig2}
\end{figure}

In the second experiment, we utilize the Normalized Mean Square Error (NMSE) as a performance metric, which is defined as
\begin{equation}
\mathrm{NMSE}\triangleq20\log_{10}(\frac{||\Xb-\hat{\Xb}||_2}{||\Xb||_2}),
\end{equation}
where $\hat{\Xb}$ is the estimate of the true signal $\Xb$. All the NMSEs are averaged over 50 Monte Carlo (MC) simulations. The number of training signals vary between $T=100$ and $T=1000$. The number of sign measurements is again $n=100$. Figure \ref{fig2} shows the NMSE performance versus the number of training signals for DL-BIHT-L2, DL-BIHT-L1 and without dictionary learning (DL) algorithm. It is seen that when $T=1000$, dictionary learning algorithms outperform the case of without dictionary learning by 4 dB performance gain. It is also observed that the proposed DL-BIHT-L2 performs slightly better than DL-BIHT-L1 and the NMSE decreases as the number of training signals increases.

\begin{figure}[tb]
\begin{center}
\includegraphics[width=8cm]{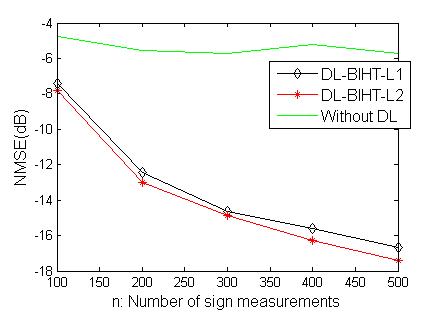}
\end{center}
\caption{NMSE versus number of sign measurements.}
%\end{center}
\label{fig3}
\end{figure}

In the third experiment, we explore the role of the number of measurements. In this case, the number of training signals is selected as $T=500$. The other parameters are the same as the second experiment. The number of sign measurements $n$ varies from 100 to 500. Figure \ref{fig3} shows the NMSE performance versus the number of sign measurements. The figure shows that with increasing the number of measurements, the performance of recovering the original signal $\Xb$ by the proposed algorithms improves. Also, both DL-BIHT-L2 and DL-BIHT-L1 significantly outperform the case of without dictionary learning algorithm. Particularly, when the number of measurements is 500, both algorithms achieve about 10 dB performance gain.

\section{Conclusion}
\label{sec: con}
We have proposed a new iterative dictionary learning algorithm for the noisy sparse signal reconstruction in one bit compressed sensing framework when the sparse domain is unknown in advance. The algorithm has two steps. The first step is the sparse signal recovery from one bit measurements which is performed by BIHT algorithm in this paper. The second step is to update the dictionary matrix. This is carried out by minimizing a suitable cost function in the one bit compressed sensing framework. A simple steepest-descent method is used to update the rows of the dictionary matrix. Simulation results show the effectiveness of the dictionary learning in monotone converging of the cost function and estimating the original signals specially when the number of training signals and the number of sign measurements increases.

%A Bayesian hypothesis test is proposed to detect the active elements of a sparse vector in one bit compressed sensing framework. Then, the amplitudes of active elements is obtained by an ML estimator. Simulation results in a special case, show that using new algorithm improves the accuracy of sparse vector estimation by at least 4dB.


\appendices
\section{Proof of the Convexity of \ltwo Norm Cost Function}
\label{sec: app1}
To verify the convexity of $J(\Db)=\sum_{i=1}^T||\yb_i-\mathrm{S}(\Db\sbb_i)||_2^2$, where $\mathrm{S}(x)$ is defined in (\ref{eq: S}), we prove that the second derivative $\frac{\partial^2 J(\Db)}{\partial^2 d_{jk}}$ is positive. First, we consider the
first order vector derivative $\frac{\partial J(\Db)}{\partial \db_j}$. Some simple calculations show that
\begin{equation}
\frac{\partial J(\Db)}{\partial \db_j}=\sum_{i=1}^T [\frac{\partial}{\partial \db_j}(-2\yb^T_i\mathrm{S}(\Db\sbb_i)+\frac{\partial}{\partial \db_j}(\mathrm{S}^T(\Db\sbb_i)\mathrm{S}(\Db\sbb_i) )].
\end{equation}
Following some other manipulations, we reach to
\begin{equation}
\frac{\partial J(\Db)}{\partial \db_j}=\sum_{i=1}^T 2\mathrm{S}'(\db^T_j\sbb_i)[-y_{ij}+\mathrm{S}(\db^T_j\sbb_i)]\sbb_i.
\end{equation}
Therefore, the scalar partial derivative $\frac{\partial J(\Db)}{\partial d_{jk}}$ is equal to
\begin{equation}
\sum_{i=1}^T[-2\mathrm{S}'(\db^T_j\sbb_i)s_{ik}y_{ij}+2\mathrm{S}'(\db^T_j\sbb_i)s_{ik}\mathrm{S}(\db^T_j\sbb_i)].
\end{equation}
The second order derivative $\frac{\partial^2 J(\Db)}{\partial d^2_{jk}}$ is
\begin{equation}
\label{eq: mid}
\sum_{i=1}^T[-2s_{ik}y_{ij}\frac{\partial}{\partial d_{jk}}(\mathrm{S}'(\db^T_j\sbb_i))+2s_{ik}\frac{\partial}{\partial d_{jk}}(\mathrm{S}'(\db^T_j\sbb_i)\mathrm{S}(\db^T_j\sbb_i))]
\end{equation}
The two partial derivatives in (\ref{eq: mid}) are equal to $\frac{\partial}{\partial d_{jk}}(\mathrm{S}'(\db^T_j\sbb_i))=\mathrm{S}''(\db^T_j\sbb_i)s_{ik}$ and $\frac{\partial}{\partial d_{jk}}(\mathrm{S}'(\db^T_j\sbb_i)\mathrm{S}(\db^T_j\sbb_i))=s_{ik}\mathrm{S}''(\db^T_j\sbb_i)\mathrm{S}(\db^T_j\sbb_i)+s_{ik}(\mathrm{S}'(\db^T_j\sbb_i))^2$.
Replacing these two terms in (\ref{eq: mid}) results in
\begin{equation}
\label{eq: fin}
\frac{\partial^2 J(\Db)}{\partial d^2_{jk}}=\sum_{i=1}^T 2s^2_{ik}[\mathrm{S}''(\db^T_j\sbb_i)(-y_{ij}+\mathrm{S}(\db^T_j\sbb_i))+(\mathrm{S}'(\db^T_j\sbb_i))^2].
\end{equation}
Consider $\mathrm{S}(x)=\frac{1-\mathrm{exp}(-x)}{1+\mathrm{exp}(-x)}$, $\mathrm{S}'(x)=\frac{2\mathrm{exp}(-x)}{(1+\mathrm{exp}(-x))^2}$ and $\mathrm{S}''(x)=\frac{-2\mathrm{exp}(-x)(1-\mathrm{exp}(-x))}{(1+\mathrm{exp}(-x))^3}$. It can be shown that for the two cases $y_{ij}=1$ and $y_{ij}=-1$, the expression in the summation in (\ref{eq: fin}) is positive. For example, consider the case $y_{ij}=1$ . Defining $x=\db^T_j\sbb_i$, with some calculations, we have
\begin{equation}
\mathrm{S}''(\db^T_j\sbb_i)(-y_{ij}+\mathrm{S}(\db^T_j\sbb_i))+(\mathrm{S}'(\db^T_j\sbb_i))^2=\frac{4\mathrm{exp}(-3x)}{(1+\mathrm{exp}(-x))^4}>0
\end{equation}
Now, consider the case $y_{ij}=-1$. In this case, it can be shown that
\begin{equation}
\mathrm{S}''(\db^T_j\sbb_i)(-y_{ij}+\mathrm{S}(\db^T_j\sbb_i))+(\mathrm{S}'(\db^T_j\sbb_i))^2=\frac{4\mathrm{exp}(-x)}{(1+\mathrm{exp}(-x))^4}>0
\end{equation}
Therefore, by proving $\frac{\partial^2 J(\Db)}{\partial d^2_{jk}}>0$, the proof of the convexity of $J(\Db)$ is complete.

\section{Proof of the Convexity of \lone-Norm Cost Function}
\label{sec: app2}
To prove the convexity of $Q(\Db)=\sum_{i=1}^T||\yb_i-\mathrm{S}(\Db\sbb_i)||_1$, where $\mathrm{S}(x)$ is given in (\ref{eq: S}), we prove that each of the sub-optimization problems $q(\db_k)=\sum_{i=1}^T||y_{ik}-\mathrm{S}(\db_k^T\sbb_i)||_1$ for $1\le k\le n$ is convex. Let $f(\db_k) = y_{ik}-\mathrm{S}(\db_k^T\sbb_i)$, then 
$\frac{\partial f(\db_k)}{\partial \db_k} = -\mathrm{S}'(\db^T_k\sbb_i)\sbb_i$ and $\frac{\partial^2 f(\db_k)}{\partial^2 \db_{k}} = -\mathrm{S}''(\db^T_k\sbb_i)\sbb^T_i\sbb_i$ are
the first and second order vector derivative of $f(\db_k)$ with respect to $\db_k$, respectively. Hence, if $\mathrm{S}''(\db^T_k\sbb_i)>0$ then $f(\db_k)$ is concave. Conversely, if 
$\mathrm{S}''(\db^T_k\sbb_i)<0$ then $f(\db_k)$ is convex. Let $z = \db^T_k\sbb_i$, then if $\mathrm{S}''(z)>0$, we have $z<0$ and as a result $y_{ik}=-1$. Hence, 
$f(\db_k)=y_{ik}-\mathrm{S}(z)=-1-\mathrm{S}(z)$ is negative. Using the composition property (\cite{Boyd04}, p. 84), since $\left \| \cdot  \right \|_1$ is convex and non-increasing when
$f(\db_k)<0$, also $f(\db_k)$ is concave, we conclude that $\left \| f(\db_k)\right \|_1$ is convex. As sum of convex functions is convex, thus $q(\db_k)$ is convex and finally $Q(\Db)$ is convex. If $\mathrm{S}''(z)<0$, we have $z>0$ and as a result $y_{ik}=1$. Hence, 
$f(\db_k)=y_{ik}-\mathrm{S}(z)=1-\mathrm{S}(z)$ is positive. Using the composition property (\cite{Boyd04}, p. 84), since $\left \| \cdot  \right \|_1$ is convex and non-decreasing when
$f(\db_k)>0$, also $f(\db_k)$ is convex, we conclude that $\left \| f(\db_k)\right \|_1$ is convex. Again, because sum of convex functions is convex $q(\db_k)$ is convex, which results in the convexity of the cost function $Q(\Db)$.

%\section*{Acknowledgment}

\ifCLASSOPTIONcaptionsoff
  \newpage
\fi

\end{document}